\documentclass{article}
\usepackage{ijcai25}

\usepackage{times}
\usepackage{soul}
\usepackage{url}
\usepackage[hidelinks]{hyperref}
\usepackage[utf8]{inputenc}
\usepackage[small]{caption}
\usepackage{graphicx}
\usepackage{amsmath}
\usepackage{amsthm}
\usepackage{booktabs}
\usepackage{algorithm}
\usepackage{algorithmic}
\usepackage[switch]{lineno}
\usepackage{amsfonts}
\usepackage{amsmath}
\usepackage{amssymb}
\usepackage{cleveref}
\usepackage{subcaption}
\usepackage{graphicx}
\usepackage{caption}
\usepackage{algorithm}
\usepackage{algorithmic}
\usepackage{amsthm}
\usepackage{pifont}

\usepackage{color}
\usepackage[dvipsnames]{xcolor}

\newcommand{\indep}{\perp \!\!\! \perp}

\usepackage{newfloat}
\usepackage{listings}
\usepackage{booktabs}
\usepackage{threeparttable}

\newtheorem{lemma}{Lemma}
\newtheorem{theorem}{Theorem}
\newtheorem{assumption}{Assumption}
\newtheorem{prop}{Proposition}

\allowdisplaybreaks[1]

\title{Optimal Policy Adaptation Under Covariate Shift}

\author{
Xueqing Liu$^1$ \and
Qinwei Yang$^1$ \and 
Zhaoqing Tian$^{1}$ \and
Ruocheng Guo$^2$ \and 
Peng Wu\footnote{corresponding author}  
\affiliations 
$^1$Beijing Technology and Business University \\
$^2$ByteDance Research \\
\emails
pengwu@btbu.edu.cn
}

\begin{document}

\maketitle
\begin{abstract}
% \lxq{under covariate shift and concept shift}
  Transfer learning of prediction models under covariate shift has been extensively studied, while the corresponding policy learning approaches are rarely discussed. In this paper, we propose principled approaches for learning the optimal policy in the target domain by leveraging two datasets: one with full information from the source domain and the other from the target domain with only covariates. First, in the setting of covariate shift, we formulate the problem from a perspective of causality and present the identifiability assumptions for the reward induced by a given policy. Then, we derive the efficient influence function and the semiparametric efficiency bound for the reward. Based on this, we construct a doubly robust and semiparametric efficient estimator for the reward and then learn the optimal policy by optimizing the estimated reward. Moreover, we theoretically analyze the bias and the generalization error bound for the learned policy. 
 Furthermore, in the presence of both covariate and concept shifts, we propose a novel sensitivity analysis method to evaluate the robustness of the proposed policy learning approach. Extensive experiments demonstrate that the approach not only estimates the reward more accurately but also yields a policy that closely approximates the theoretically optimal policy.  
% %  but generalize to the entire domain 
  %\lxq{Furthermore, our proposed approach achieves superior performance consistently under different degree of covariate shift and concept shift. Extensive experiments demonstrate that the approach not only estimates the reward more accurately but also yields a policy that closely approximates the theoretically optimal policy. }  
  %对不同的偏移程度都保持稳定。 
\end{abstract}

\section{Introduction}
In many real-world scenarios, labeled data is often scarce due to budget constraints and time-consuming collection processes ~\cite{zhuang2020comprehensive,imbens2024long}, significantly limiting the generalizability of the resulting models. For example, in medical research, collecting labeled data involves extensive clinical trials and follow-up periods, making it costly and time-consuming ~\cite{Dahabreh-etal2020,hu2023longterm}. In autonomous driving, obtaining labeled data requires manual annotation of large amounts of sensor data, which is laborious and expensive ~\cite{Sun_2020_CVPR}. 
To address this problem and enhance a model's performance in a target domain without labels, an active area of research is transfer learning. 
It aims to improve the performance of target learners in the target domain by transferring the knowledge contained in a different but related source domain. 
While transfer learning has been extensively studied in the context of prediction models \cite{wang2018deep,wang2020novel,pesciullesi2020transfer}, how to transfer a policy is still underdeveloped. Policy learning refers to identifying individuals who should receive treatment/intervention based on their characteristics by maximizing rewards~\cite{murphy2003optimal}. 
It has broad applications in recommender systems \cite{chen2021knowledge,Wu-etal2022}, precision medicine \cite{bertsimas2017personalized} and reinforcement learning \cite{liu2021policy,kwan2023survey}. 
Unlike transfer learning for prediction models, policy transfer faces identification challenges due to its counterfactual nature ~\cite{athey2021policy,li2023trustworthy,Wu-etal-ShortLong,yang2024learning}. Instead of predicting outcomes based on observed data, policy transfer requires considering what would happen under different actions, making the process more complex.  % directly 
%  the target data has covariates 
%  of policies
% Unlike transfer learning for prediction models, policy transfer faces identification challenges due to its counterfactual nature. Instead of predicting outcomes from observed data, policy transfer must account for potential outcomes under different actions, adding complexity to the process.

% In contrast to the transfer learning of predictive models, policy transferring encounters problems of identification due to its feature of counterfactual. As there is significant difference between source data and target data, covariate shift is common. The shift in the covariate/predictor distribution, often referred to as covariate shift, is one of the key contributors to poor transportability and generalizability of a supervised learning model from one data set to another~\cite{liu2023augmented}. 

%Several works consider the estimation of parameter model in the setting of semi-supervised learning. However, how to transfer the policy remains unaddressed.

% by using a dataset from source domain and a dataset with only covariates from the target domain.    
% In this article, 
%We assume that the source and target datasets are independent, with
We aim to learn the optimal policies in the target and entire domains using a dataset from the source domain (source dataset) and a dataset from the target domain (target dataset).
% which are applied to the target domain and the wholedomain using a dataset from the source domain and a dataset from the target domain.  
The source dataset includes the covariates, treatment, and outcome for each individual, whereas the target dataset contains only the covariates. 
We assume that the source dataset satisfies the unconfoundedness and overlap assumptions while imposing fewer restrictions on the target dataset.
We allow for substantial differences in the covariate distributions between the source and target datasets (referred to as covariate shift), while assuming that the conditional distributions of potential outcomes given covariates are the same.
In this article, we first propose a principled policy learning approach under covariate shift.  
Specifically, we define the reward and the optimal policy in the target domain using the potential outcome framework in causal inference. 
Under the widely used assumptions of unconfoundedness and transportability, we establish the identifiability of the reward in the target domain and then derive its efficient influence function and semiparametric efficiency bound. Building on this, we develop a novel estimator for the reward. Theoretical analysis shows that the proposed estimator is doubly robust and achieves the semiparametric efficient bound, that is, it is the optimal regular estimator in terms of asymptotic variance ~\cite{newey1990semiparametric}.
% Additionally, the proposed estimator \pw{enjoys} the property of double robustness.       
% that is shown to be semi-parametric efficient, i.e., it is a optimal regular estimator in terms of asymptotic variance \cite{tsiatis2006semiparametric}.
% of the reward in target domain
Then we propose to learn the optimal policy by maximizing the estimated reward. We analyze the bias of the estimated reward and the generalization error bound of the learned policy.  
In addition, we extend the proposed method to learn the optimal policy in the entire domain consisting of the source and target domains by leveraging data from both domains to address distributional discrepancies and ensure robust generalization across heterogeneous environments.
  % Furthermore, considering that the transportability assumption, i.e., the conditional distributions of potential outcomes given covariates remain unchanged, may be violated under concept shift, we propose a novel sensitivity analysis method for concept shift. This method evaluates the robustness of the approach under varying degrees of concept shift, accounting for the presence of both covariate and concept shifts. 
  
The main contributions are summarized as follows: (1) We propose a principled approach for learning the optimal policy under covariate shift from a perspective of causality, by introducing plausible identifiability assumptions and efficient estimation methods; (2) We provide a comprehensive theoretical analysis of the proposed approach, including the consistency, asymptotic normality, and semiparametric efficiency of the estimator of reward. Additionally, we derive the bias and generalization error bound for the learned policy; (3) We conduct extensive experiments to demonstrate the effectiveness of the proposed policy learning approach.

\section{Problem Formulation} 
%In this section, 
%We formulate the problem using the potential outcome framework in causal inference ~\cite{splawa1990application}.  
% \col{from a source domain to a target domain} ,
% of transferring policies 

\subsection{Notation and Setup} 
Let $A \in \mathcal{A} = {\{0, 1\}}$ denote the binary indicator for treatment, where $A = 1$ indicates receiving treatment and $A = 0$ indicates not receiving treatment. The random vector $X \in \mathcal{X} \subset \mathbb{R}^p$ represents the $p$-dimensional covariates measured before treatment, and $Y \in \mathcal{Y} \subset \mathbb{R}$ denotes the outcome of interest. Assume that a larger outcome is preferable. 
Under the potential outcome framework~\cite{rubin1974estimating,splawa1990application}, let $Y(a)$ denote the potential outcome that would be observed if $A$ were set to $a$ for $a \in \mathcal{A}$. By the consistency~\cite{Hernan-Robins2020}, the observed outcome $Y$ satisfies $Y = Y(A) = AY(1) + (1 - A)Y(0)$.

%Suppose that we have access to
% Let $G\in \mathcal{G} \subset \mathbb{R}$ be the indicator for data source, with $G=1$ for the randomized controlled trial (RCT) and $G=0$ for observational study (OBS). The random vector $X \in \mathcal{X} \subset \mathbb{R}^p$ represents a $p$-dimensional pre-treatment covariates, and $Y \in \mathcal{Y} \subset \mathbb{R}$ denotes the outcome of interest. Under the potential outcome framework \cite{rubin1974estimating,splawa1990application}, let $Y(a)$ denote potential outcome that would be observed if $A$ has been set to $a$ for $a\in\{0,1\}$. By the consistency assumption, the observed outcome $Y$ is $Y = Y(A) = AY(1) + (1 - A)Y(0)$.
%设定数据集D，D1,D2,Y(A),n
% By the design, the observed data consists of a RCT dataset and a OBS dataset, denoted as  
Without loss of generality, we consider a typical scenario involving two datasets: a source dataset and a target dataset, which are representative samples of the source domain and target domain, respectively. Let $G \in  \{ 0, 1\}$ be the indicator for the data source, where $G = 1$ denotes the source domain and $G = 0$ denotes the target domain. The observed data are represented as follows,  
   \begin{align*}
\mathcal{D}_1 ={}&  \{ (X_i,A_i,Y_i,G_i=1): i = 1,...,n_1 \}, \\
\mathcal{D}_0 ={}&  \{ (X_i,G_i=0): i = n_1 + 1,...,n_1 + n_0 \}, 
   \end{align*}
   where the source dataset $\mathcal{D}_1$ consists of $n_1$ individuals, with observed covariates, treatment, and outcome for each individual.
%\begin{table}[ht!]
%\centering
%\caption{Observed data structure. "\checkmark" and "\xmark" mean observed and unobserved, respectively}
%%\resizebox{.95\columnwidth}{!}{
%\begin{tabular}{cccccc} 
%\toprule %表格顶部粗横线
%Dataset & Unit & G & X & A & Y \\
%\midrule % 表格中间细横线
%  & 1 & 1 & \checkmark & \checkmark & \checkmark \\
%$\mathcal{D}_1$ & … & … & … & … & … \\
%  & $n_1$ & 1 & \checkmark & \checkmark & \checkmark \\
%\midrule % 表格中间细横线
%  & $n_1 + 1$ & 0 & \checkmark & \xmark & \xmark \\
%$\mathcal{D}_0$ & … & … & … & … & … \\
%  & $n = n_1 + n_0$ & 0 & \checkmark & \xmark & \xmark \\
%\bottomrule % 表格底部粗横线
%\end{tabular}
%\label{table1}
%%\vspace{-1.5em}---不能用
%\end{table}
%where $n_1$ and $n_0$ are the sample size of RCT data and OBS data, respectively. 
% where the source dataset $\mathcal{D}_1$ has $n_1$ individuals, and for each individual, we have the full information on covariates, treatments, and outcomes.   
% The target dataset  $\mathcal{D}_0$ contains $n_0$ individual, for which only covariates are available for each individual. 
% Let $n = n_0 + n_1$. Table \ref{table1} illustrates the observed data structure.  
 The target dataset $\mathcal{D}_0$ contains $n_0$ individuals, with only covariates for each individual. 
% where only covariates are available for each individual.
This is common in real life due to the scarcity of outcome data. For example, in medical research, patient features are observed, but obtaining outcomes requires long-term follow-up~\cite{hu2023longterm,imbens2024long}. Let $\mathbb{P}(\cdot| G=1)$ and $\mathbb{P}(\cdot| G=0)$ be the distributions of the two datasets respectively.
Then $n = n_0 + n_1$ and $q=n_1/n$ represent the
probability of an individual belonging to the source population. 

\subsection{Formulation}
%给出在RCT里reward的表达式

% It should be noted that based only the target dataset, we cannot 

% Therefore we explore how to borrow strength from source dataset in target dataset. 

% Now we give formalization about learning an optimal policy in the observational study. 
%of learning the optimal policy in the target domain
We formulate the goal of learning the optimal policy in the target domain. Specifically, let $\pi: \mathcal{X} \to \mathcal{A}$ denote a policy that maps individual covariates  $X = x$ to the treatment space $\mathcal{A}$. A policy $\pi(X)$ is a treatment rule that determines whether an individual receives treatment $(A = 1)$ or not $(A = 0)$. 
% We formulate the goal of learning the optimal policy in the target domain. Let $\pi: \mathcal{X}\to \mathcal{A}$ denote a policy that maps individual context $X = x$ % in $\mathcal{D}_0$ 
% to the treatment space $\mathcal{A} = \{0, 1\}$. A policy $\pi(X)$ is a treatment rule that determines whether an individual receives treatment $(A = 1)$ or not $(A = 0)$. 
%\col{In observational studies, the choice of treatment allocation can have significant consequences for outcomes of interest, such as health outcomes in medical studies or user engagement metrics in online experiments. Therefore, the policy $\pi(X)$ plays a crucial role in determining the effectiveness and fairness of the interventions applied to different individuals.} 
For a given policy $\pi$ applied to the target domain, the average reward is defined as follows % (\pw{also policy value})  
\begin{equation}\label{eq1}
R(\pi)=\mathbb{E}[\pi(X)Y(1) + (1-\pi(X))Y(0) | G = 0]. 
\end{equation} 
We aim to learn the optimal policy $\pi^*$ defined by  
$\pi^* =  \arg \max_{\pi\in\Pi} R(\pi)$, 
%\begin{equation*}
%\pi^* =  \arg \max_{\pi\in\Pi} R(\pi)
%\end{equation*}
where $\Pi$ is a pre-specified policy class. For example, $\pi(X)$ can be modeled with a parameter $\theta$ using methods such as logistic regression or multilayer perceptron, with each value of $\theta$ corresponding to a different policy.
%such as logistic regression and multilayer perceptron (MLP). 
%除此之外，根据对一个给定的作用在两个领域上的\pi(x)，平均reward定义为V，V的具体细节在第5章会探究

In addition, for a policy $\pi(x)$ applied across the whole domain, the corresponding average reward is defined as 
% $V(\pi)=\mathbb{E}[\pi(X)Y(1) + (1-\pi(X))Y(0)].$  
\begin{equation}\label{eq2}
V(\pi)=\mathbb{E}[\pi(X)Y(1) + (1-\pi(X))Y(0)]. 
\end{equation} 

There is a subtle difference between $R(\pi)$ and $V(\pi)$. 
 For $R(\pi)$, our focus is on transferring the policy from the source domain to the target domain, and for $V(\pi)$, we aim to generalize the policy from the source domain to the entire domain. 
In the main text, we focus on learning the policy maximizing $R(\pi)$ to avoid redundancy. 
We also develop a similar approach to learn the policy maximizing $V(\pi)$ and briefly present it in Section \ref{sec4-3}.
\section{Oracle Policy and Identifiability} \label{sec3}

\subsection{Oracle Policy}
The optimal policy that maximizes Eq. \eqref{eq1} has an explicit form. Let $\tau(X) = \mathbb{E}[Y(1) - Y(0) | X, G=0]$ be the conditional average treatment effect (CATE) in the target domain, 
% then   
\begin{equation*}\begin{aligned}
R(\pi)&=\mathbb{E}[\pi(X)\{Y(1)-Y(0)\} + Y(0) | G=0]\\
&=\mathbb{E}[\pi(X)\tau(X) | G=0] + \mathbb{E}[Y(0) | G=0]
\end{aligned}\end{equation*}
where the last equality follows from the law of iterated expectations. Then we have the following conclusion. 
%the oracle policy $\pi_{0}^{*}(x)$ can be represented by Lemma \ref{lemma1}.
\begin{lemma}
\label{lemma1}
The oracle policy
\begin{equation*}\begin{aligned}
\pi_{0}^{*}(x)& =\arg\max_{\pi}R(\pi) 
%=\arg\max_{\pi}\mathbb{E}[\pi(X)\tau(X) | G = 0] \\
=\begin{cases}1,&\text{if }\tau(x)\geq0\\0,&\text{if }\tau(x)<0,\end{cases}
\end{aligned}\end{equation*}
where % $ \mathbb{E}[Y(0) | G=0]$ is a constant and 
$\max_{\pi}$ is taken over all possible policies without constraints, rather than being restricted to $\Pi$.   
\end{lemma}

For an individual characterized by $X = x$ in the target domain, Lemma \ref{lemma1} asserts that the decision to accept treatment ($A = 1$) should be based on the sign of $\tau(x)$. The oracle policy $\pi_{0}^{*}$ recommends treatment for individuals expected to experience a positive benefit, thereby optimizing the overall reward within the target domain.  The target policy $\pi^*$ equals the oracle policy $\pi^*_0$ in Lemma \ref{lemma1} if $\pi^*_0 \in \Pi$; otherwise, they may not be equal, and their difference is the systematic error induced by limited hypothesis space of $\Pi$.   
% For an individual characterized by $X=x$,  Lemma \ref{lemma1} asserts that the decision to accept treatment ($A = 1$) should be based on the sign of $\tau(x)$. The oracle policy $\pi_{0}^{*}$  
% recommends treatment for individuals that are expected to have a positive benefit, thereby optimizing the overall reward within the target population. 
%This decision rule is critical for ensuring that treatments are administered only when they are expected to provide a positive benefit, thereby optimizing the overall outcomes within the study.
% If $\tau(x)$ is positive, indicating a beneficial effect of the treatment for the unit with covariates $x$, the optimal policy advises accepting the treatment. Conversely, if  $\tau(x)$ is negative or zero, suggesting either a detrimental effect or no effect, the policy recommends against accepting the treatment($A = 0$). This decision rule is critical for ensuring that treatments are administered only when they are expected to provide a positive benefit, thereby optimizing the overall outcomes within the study.

\subsection{Identifiability of the Reward}
% Since the target dataset only contains covariates $X$, $R(\pi)$ cannot be identified based on the target data alone due to the lack of treatment and outcome information.  To recover identifiability, it is necessary to borrow information from the source dataset by imposing several assumptions. 
To learn the optimal policy $\pi^*$, we first need to address the identifiability problem of $R(\pi)$, as this forms the foundation for policy evaluation. Since the target dataset only contains covariates $X$, $R(\pi)$ cannot be identified from the target data alone due to the absence of treatment and outcome. To identify $R(\pi)$, it is necessary to borrow information from the source dataset by imposing several assumptions. 
 % The identifiability   
 % and impose
 % to identify reward  $R(\pi)$ by presenting the following assumptions. 
%and both the outcome and the treatment are unobserved, it is necessary to borrow strength from RCT dataset to identify reward  $R(\pi)$ by presenting the following assumptions.

% [Strong Ignorability]
\begin{assumption} \label{assumption1}  For all $X$ in the source domain,  

(i) Unconfoundedness: $(Y(1), Y(0)) \indep A \mid X, G=1$;

(ii) Overlap: $0\textless e_1(X) \triangleq \mathbb{P}(A = 1 | X, G = 1)  \textless 1$, where $e_1(X)$ is the propensity score. %~\cite{Rosenbaum-Rubin-1983}. %  for all  $X$
\end{assumption}
%The $e_1(X)$ is the propensity score. 
% thereby excluding unmeasured confounding.  %  in the source domain 

% Assumption \ref{assumption1}(i) states that, in the source domain, the treatment $A$ is independent of the potential outcomes $(Y(1), Y(0))$ given thecovariates $X$, 
Assumption \ref{assumption1}(i) states that, in the source domain, the treatment is independent of the potential outcomes given the
covariates,  
implying that all confounders affecting both the treatment and outcome are observed.   
Assumption \ref{assumption1}(ii) asserts that any individual characterized by $X$ in the source domain has a positive probability of receiving treatment.  
Assumption \ref{assumption1} is a standard assumption for identifying causal effects in the source domain~\cite{Rosenbaum-Rubin-1983}. 
%is positive for any given set of covariates.
% Assumption \ref{assumption2}(ii) asserts that the probability of an individual receiving treatment for any given covariates is positive in the source dataset.
%\col{holding true for randomized controlled trials or unconfounded observational studies.} 
However, Assumption \ref{assumption1} is not enough to identify the causal effects in the target domain. Thus, we further invoke  Assumption \ref{assumption2}. %  below.  
% Assumption \ref{assumption1} can only identify causal effects in the source domain. Therefore, we present Assumption \ref{assumption2} to further identify the causal effects in the target dataset.  

\begin{assumption}[Transportability]  \label{assumption2}
Suppose that 

(i) $(Y(0),Y(1)) \indep G \mid X$ for all $X$;

(ii) $0  \textless {s(X)} \triangleq \mathbb{P}(G = 1 | X)  \textless 1$ for all $X$ in the source domain, where $s(X)$ is the sampling score.  
% $\mathbb{E}[Y(0) | X, G=0] = \mathbb{E}[Y(0) | X, G=1]$

\end{assumption}

% (i) is a standard assumption and wildly used in estimating causal effects via data combination in causal inference~\cite{Hotz-etal2005, Cole-Stuart2010, stuart2011use, Tipton2013, Hartman-etal2015, Kern2016, lesko2017generalizing, Buchanan-etal2018, Li-etal2022, Li-etal2023}. 
Assumption \ref{assumption2}  is widely adopted in causal effects estimation via data combination in causal inference ~\cite{stuart2011use,hartman2015sample,Kern2016,lesko2017generalizing,Buchanan-etal2018,Li-etal2022,li2023balancing,colnet2024causal,Wu-etal-2025-Compare,wu-mao2025}. %Tipton2013,
Assumption \ref{assumption2}(ii) indicates that all individuals in the source domain have a positive probability of belonging to the target domain. Assumption \ref{assumption2}(i) implies that $\mathbb{E}[Y(a) | X, G=1] = \mathbb{E}[Y(a) | X, G=0] = \mathbb{E}[Y(a) | X ]$ for $a = 0, 1$, which ensures the transportability of the CATE from the source domain to the target domain and leads to the identifiability of $\tau(X)$, that is, 
% Note that   
%  \[     \frac{\mathbb{P}(G=1|X) }{ \mathbb{P}(G=0|X) } =  \frac{ \mathbb{P}(X| G=1) \mathbb{P}(G=1) }{ \mathbb{P}(X| G=0) \mathbb{P}(G=0) },    \]
% Assumption \ref{assumption2}(ii) also implies that the support of the covariates in the target domain must be larger than that in the source domain. Otherwise, there could be cases where $\mathbb{P}(X|G=1) > 0$ while $\mathbb{P}(X|G=0) = 0$, leading to a violation of Assumption \ref{assumption2}(ii). 
%which would violate 
% leading to a violation of   
 % Assumption \ref{assumption2}(i) implies  that   $\mathbb{E}[Y(a) | X, G=1] = \mathbb{E}[Y(a) | X, G=0]$ for $a = 0, 1$,   
%\begin{equation*}  
%\mathbb{E}[Y(a) \mid X, G=1] = \mathbb{E}[Y(a) \mid X, G=0].
%\end{equation*}  
 % which ensures the transportability of the CATE from the source domain to the target domain. 
 %Then under Assumptions \ref{assumption1}--\ref{assumption2}, 
 % we have
\begin{align*} %\label{eq-tmp}
 \tau&(X) 
 %= \mathbb{E}[Y(1) - Y(0) | X,G=0] \\
 =  \mathbb{E}[Y(1) - Y(0) | X,G=1]\\
 &=  \mathbb{E}[Y|X,A=1,G=1] - \mathbb{E}[Y|X,A=0,G=1]    \\
 &\triangleq  \mu_{1}(X) - \mu_{0}(X),
\end{align*} 
where the third equality follows from Assumption \ref{assumption1}. 
% Eq. (\ref{eq-tmp}) leads to the identifiability of $\tau(X)$, i.e., the CATE in the target domain.  
%As a consequence, under Assumption 1-2,
Thus, under Assumptions \ref{assumption1}-\ref{assumption2}, the reward $R(\pi)$ can be identified as 
\begin{align}
\label{eq4}
&R(\pi)={} \mathbb{E}[\pi(X)\mu_{1}(X) + (1-\pi(X)) \mu_{0}(X)| G=0] \notag \\
={}& \mathbb{E}\Big[ \frac{1-G}{1-q} \{ \pi(X)\mu_{1}(X) + (1-\pi(X)) \mu_{0}(X)\}  \Big]
\end{align}

Assumption \ref{assumption2} allows the presence of covariate shift, i.e., the distribution of $X$ in the source domain may significantly differ from that in the target domain~\cite{gama2014survey}. 

\section{Policy Adaptation Under Covariate Shift} \label{sec4}

% 只基于RCT数据不能够得到OBS上的；只基于OBS数据也不行；
% develop a principled approach for 
In this section, we proposed a method for learning the optimal policy. It consists of two steps: (a) policy evaluation, estimating the reward $R(\pi)$ for a given $\pi$, and (b) policy learning, learning the optimal policy based on the estimated reward.   % policy 
% The proposed method 
% The proposed method
%  Incorporating RCT Data
\subsection{Estimation of the Reward $R(\pi)$} \label{sec4-1}
% 

% \bcol{[Polish: Zhaoqing Tian]}
% To improve the accuracy of estimating the rewards $R(\pi)$, we first give the following two debiasing methods. 
%无偏性，
% For estimating the reward $R(\pi)$, 
According to Eq. (\ref{eq4}), a direct method for estimating the reward $R(\pi)$ is given as
\begin{align*}
       \hat{R}_{\text{Direct}}&(\pi) ={} \frac{1}{n} \sum_{i = 1}^{n}   \frac{1-G_i}{1-q}  \\
      &{} \times \left \{  \pi(X_i) \hat \mu_1(X_i)  + (1 -  \pi(X_i)) \hat \mu_0(X_i) \right \},
 \end{align*}
 where $\hat \mu_a(X)$ ($a= 0, 1$) represents the estimated outcome regression function $\mu_a(X)$.
 %defined in Eq. (\ref{eq-tmp}). 
 This can be implemented by regressing $Y$ on $X$ using the source dataset with $A=a$.  
%The consistency\footnote{An estimator is consistent if it converges to the true value of the parameter in probability as the sample size increases} (asymptotic unbiasedness) 
 The unbiasedness of the direct estimator $\hat{R}_{\text{Direct}}(\pi)$ depends on the accuracy of $\hat \mu_a(X)$. 
 % the estimated outcome regression functions. 
 If $\hat \mu_a(X)$ is a biased estimator of $\mu_a(X)$, then $\hat{R}_{\text{Direct}}(\pi)$ will also be a biased estimator of $R(\pi)$.  
  % In addition, 
% As discussed in other literature,
%From a machine learning perspective, 
Moreover, the generalization performance of the direct method is often poor because $\hat \mu_a(X)$ is trained using the source dataset of $A=a$, but is applied to the whole target dataset~\cite{Li2022TDRCLTD,Wu-Han2024}. When there is a significant difference in the covariate distributions between the source and target datasets, the direct method suffers from the problem of model extrapolation, resulting in poor practical performance.   
% ~\cite{Tan-2007}
%From a perspective of machine learning, the generalization performance is usually poor due to the truth that $\hat \mu_a(X)$ are trained using the source dataset of $A=a$, while are used in the target dataset. When there is a significant difference in the covariates distribution between the source and target domains, the direct method suffers the problem of model extrapolation, leading to poor performance in practice. 

  %difficulty in obtaining accurate outcomes.
% The EIB leverages the information effectively from OBS dataset, particularly extracting the relationship between  covariates $X$ and outcomes $Y$. 
% However, its generalization performance is poor due to the difficulty in obtaining accurate outcomes. It remains unbiased in the presence of only covariate shift, but it becomes ineffective when concept drift is also present.
% Dirct method主要利用了X与Y的关系; 
% 局限性:  当只存在covariat shift，它是无偏的，但它有泛化的问题；当存在concept drift的情况下，它失效了。

% Specifically, we n
In addition to the direct method, one can use the propensity score $e_1(X)=  \mathbb{P}(A = 1 | X, G = 1)$ and sampling score $s(X) = \mathbb{P}(G=1|X)$  to construct the inverse probability weighting (IPW) estimator of $R(\pi)$. Note that 
% \begin{equation} %  \label{eq5}
%   \begin{aligned}
  {\footnotesize\begin{align*}
             R(\pi) = \mathbb{E} & \Big [ \frac{G}{1-q} \omega(X) \left \{ \frac{\pi(X)AY}{e_1(X)} + \frac{(1-\pi(X))(1-A)Y}{1-e_1(X)} \right \}\Big ],
   \end{align*}}with $\omega(X)=(1-s(X))/s(X)$. 
  %    \end{aligned}
  % \end{equation}} 
  Based on it, the IPW estimator of $R(\pi)$ is given as  
\begin{align*}
  \hat{R}_{\text{IPW}}&(\pi) 
  ={}\frac{1}{n} \sum_{i = 1}^{n} \Big \{ \frac{G_i}{1-q}  \frac{\pi(X_i) A_i  Y_i }{\hat e_1(X_i)} \frac{1 - \hat s(X_i)}{\hat s(X_i)} \\ 
  & + \frac{G_i}{1-q}  \frac{ (1 - \pi(X_i)) (1-A_i)Y_i }{1 - \hat e_1(X_i)}  \frac{1 - \hat s(X_i)}{\hat s(X_i)}\Big \}, 
\end{align*}
where $\hat e_1(X)$ and $\hat s(X)$ are estimates of $e_1(X)$ and $s(X)$, respectively. 
% Secondly, based on Eq. (\ref{eq4}) and the introduction of the definition of the propensity score $e_1(X):=  \mathbb{P}(A = 1 | X, G = 1)$,  the inverse-propensity-scoring (IPS) methods \cite{schnabel2016recommendations}  or estimating $R(\pi)$ is given as
% 通过引入ps，可以构造IPS估计
The IPW estimator $\hat{R}_{\text{IPW}}(\pi)$ is an unbiased  estimator of $R(\pi)$ when $\hat e_1(X)$ and $\hat s(X)$ are accurate estimators of $e_1(X)$ and $s(X)$, respectively, i.e., 
$\hat e_1(X) = e_1(X)$ and $\hat s(X) = s(X)$.    
However, a limitation of the IPW estimator is its inefficiency, meaning it tends to have a large variance. 
% This will be evident in our experiments.   
%In our experiments, we will see this. 
% However the IPW estimator is highly
% sensitive to small propensity score $e_1(x)$ and $s(x)$, which leads to the high variance problem.
%\col{Limitations of IPS estimator.}
% 方差可能比较大；尤其是e_1(x)或s(x)比较小的时候。

% specifically in not fully utilizing the observed data. 
% The EIB method does not leverage the information of data sources $G$ and treatments $A$, while the IPW method does not extracting the relationship between covariates $X$ and outcomes $Y$.  
% Both of the direct and IPW methods have certain limitations, which are
The limitations of direct and IPW methods are essentially caused by the insufficiency of utilizing the observed data. The direct method does not leverage the information of data indicator $G$ and treatment $A$, while the IPW method does not extract the relationship between covariates $X$ and outcome $Y$.  
% In order 
To fully utilize the observed data, we 
employ the semiparametric efficiency theory~\cite{tsiatis2006semiparametric} to  
derive the efficient influence function and the efficiency bound of $R(\pi)$. This allows us to obtain the semiparametric efficient estimator of $R(\pi)$. A semiparametric efficient estimator is considered optimal as it reaches the semiparametric efficiency bound, meaning it has the smallest asymptotic variance under several regularity conditions~\cite{newey1990semiparametric}.   
% We aim to characterize the semiparametric efficiency bound \cite{newey1990semiparametric} of $\pi(X)$ when OBS data is available.
% The following Theorem \ref{theo1} presents efficient influence function and semiparametric efficiency bound of  $R(\pi)$ under Assumptions \ref{assumption1}-\ref{assumption2}. 
% We first present the semiparametric efficiency bounds and  efficient influence function of $R(\pi)$ to derive efficient estimators.  
\begin{theorem}[Efficiency Bound of $R(\pi)$] 
\label{theo1} Under Assumptions \ref{assumption1}-\ref{assumption2}, the efficient influence function of  $R(\pi)$ is 
{\small\begin{align*}
  &\varphi_{R} ={} \frac{G}{1-q}  \frac{\pi(X) A  \{Y - \mu_1(X)\}}{e_1(X)} \frac{1 - s(X)}{s(X)} \\ 
  & + \frac{G}{1-q}  \frac{ (1 - \pi(X)) (1-A)\{Y - \mu_0(X)\}}{1 - e_1(X)}  \frac{1 - s(X)}{s(X)} \\
 & +  \frac{1-G}{1-q} \{ \pi(X)\mu_1(X) + (1 - \pi(X)) \mu_0(X) - R(\pi) \}. 
\end{align*}}The semiparametric efficiency bound of $R(\pi)$ is $\text{Var}(\varphi_{R})$.
\end{theorem}
% (See Appendix A for proofs) 
Theorem \ref{theo1} (See Appendix A.1 for proofs)  presents the efficient influence function and semiparametric efficiency bound of $R(\pi)$ under Assumptions \ref{assumption1}-\ref{assumption2}. 
% the efficient influence function of $R(\pi)$,
%which are crucial for constructing efficient estimators of $R(\pi)$. 
From Theorem \ref{theo1}, we can construct the semiparametric efficient (SE) estimator of $R(\pi)$, which is given as 
{\small\begin{align*}
  &\hat{R}_{\text{SE}}(\pi)
  % &={}\frac{1}{n} \sum_{i=1}^n \varphi_{R(\pi)}(X_i, A_i, Y_i, G_i, q,  \pi, \hat s, \hat e_1, \hat \mu_1,\hat \mu_0)\\
  ={}\frac{1}{n} \sum_{i = 1}^{n} \Big [ \frac{G_i}{1-q}  \frac{ \pi(X_i) A_i  \{Y_i - \hat \mu_1(X_i)\}}{\hat e_1(X_i)} \frac{1 - \hat s(X_i)}{\hat s(X)} \\ 
  & + \frac{G_i}{1-q}  \frac{ (1 -  \pi(X_i)) (1-A_i)\{Y_i - \hat \mu_0(X_i)\}}{1 - \hat e_1(X_i)}  \frac{1 - \hat s(X_i)}{\hat s(X_i)} \\
 & +  \frac{1-G_i}{1-q} \{  \pi(X_i) \hat \mu_1(X_i) + (1 -  \pi(X_i)) \hat \mu_0(X_i) \} \Big ]. 
\end{align*}}
Next, we analyze the theoretical properties of $\hat{R}_{\text{SE}}(\pi)$.  % the proposed estimator 
% where $\hat s(X)$, $\hat e_1(X)$ and $\hat \mu_a(X)$ for $a = 0,1$ are estimates of  $s(X)$, $ e_1(X)$ and $ \mu_a(X)$ respectively.  
% Suppose that $\bar e_1(X)$, $\bar s(X)$, $\bar \mu_1(X)$, $\bar \mu_0(X)$ are the probability limit of $\hat e_1(X)$, $\hat s(X)$, $\hat \mu_1(X)$, $\hat \mu_0(X)$ , respectively. When a model is correctly specified, then its probability limit will equal to its true value; otherwise, they may differ. 
\begin{prop}[Double Robustness of $\hat{R}_{\text{SE}}(\pi)$]
\label{prop2}
 Under Assumptions \ref{assumption1} and \ref{assumption2}, $\hat{R}_{\text{SE}}(\pi)$ is an unbiased estimator of $R(\pi)$ if one of the following conditions is satisfied: 
 % (i)  $\hat \mu_a(x) = \mu_a(x)$, i.e., $\hat \mu_a(x)$ estimates $\mu_a(x)$ accurately for $a = 0, 1$; (ii)  $\hat e_1(x) = e_1(x)$ and $\hat s(x) = s(x)$, i.e.,  $\hat e(x)$ and $\hat s(x)$ estimate $e(x)$ and $s(x)$ accurately. 
\begin{itemize}
    \item[(i)] $\hat \mu_a(x) = \mu_a(x)$, i.e., $\hat \mu_a(x)$ estimates $\mu_a(x)$ accurately for $a = 0, 1$.     
    \item[(ii)] 
      $\hat e_1(x) = e_1(x)$ and $\hat s(x) = s(x)$, i.e.,  $\hat e(x)$ and $\hat s(x)$ estimate $e(x)$ and $s(x)$ accurately. 
\end{itemize}
\end{prop}

%  (See Appendix B for proofs) 
Proposition \ref{prop2} (See Appendix A.2 for proofs) shows the
double robustness of $\hat{R}_{\text{SE}}(\pi)$, 
i.e., if the outcome regression function $\mu_a(X)$ for $a = 0, 1$ can be estimated accurately, or the propensity score $e_1(X)$ and the sampling score the $s(X)$ can be estimated accurately, $\hat{R}_{\text{SE}}(\pi)$ is unbiased estimator of $R(\pi)$.   
Compared to the direct method that requires $\hat \mu_a(X) = \mu_a(X)$ for unbiasedness, and the IPW method that requires $\hat e_1(X) = e_1(X)$ and $\hat s(X) = s(X)$ for unbiasedness, double robustness provides more reliable results by mitigating the inductive bias caused by inaccurate models for the nuisance parameters $e_1(X), s(X),$ and $\mu_a(X)$ for $a = 0, 1$. 
% i.e., if  $\mu_a(X)$ for $a = 0, 1$ can be estimated accurately, or $e_1(X)$ and $s(X)$ can be estimated accurately, $\hat{R}_{\text{SE}}(\pi)$ is an unbiased estimator of $R(\pi)$.   
% Double robustness provides more reliable results by mitigating inductive bias caused by inaccurate models for $\{e_1(X), s(X)\} $ and $\mu_a(X)$ for $a = 0, 1$.
% nuisance parameters

%Compared to the direct method that requires $\hat \mu_a(X) = \mu_a(X)$ for unbiasedness, and the IPW method that requires $\hat e_1(X) = e_1(X)$ and $\hat s(X) = s(X)$ for unbiasedness, double robustness provides more reliable results by mitigating the inductive bias caused by inaccurate models for nuisance parameters $e_1(X), s(X),$ and $\mu_a(X)$. % for $a = 0, 1$. 

% double robustness offers more reliable results by mitigating the inductive bias introduced by inaccuracies in the models for the nuisance parameters

% The model based on direct method requires that $\mu_{a}(X)=\mathbb{E}[Y|X,A=a,G=1]$ for $a=0,1$ be correctly specified in order for the reward $\mathbb{R}(\pi)$ to be estimated without bias. In a similar manner, the model based on IPW method requires that $e_1(X):=  \mathbb{P}(A = 1 | X, G = 1)$ be correctly specified in order for the unbiased estimator of reward $\mathbb{R}(\pi)$.
% Compared it with Direct and IPW method, double robustness provides more reliable results, reducing the systematic bias caused by model misspecification.
%model misspecification.
%\col{Compare it with Direct and IPS methods}. 

%\begin{prop}[variance reduction] 
\begin{theorem}[Efficiency of $\hat{R}_{\text{SE}}(\pi)$] \label{thm2}
    Under the Assumptions \ref{assumption1}--\ref{assumption2}, if $|| \hat{e}_1(x) - e_1(x) ||_2 \cdot || \hat{\mu}_a(x) - \mu_a(x) ||_2 = o_\mathbb{P}(n^{-1/2})$ and $|| \hat{s}(x) - s(x) ||_2 \cdot || \hat{\mu}_a(x) - \mu_a(x) ||_2 = o_\mathbb{P}(n^{-1/2})$ for all $x \in \mathcal{X}$ and $ a \in \{0, \ 1 \}$,
    then $\hat{R}_{\text{SE}}(\pi)$ is a consistent estimator of $R(\pi)$, and satisfies
    \begin{equation*}
        \sqrt{n}\{ \hat{R}_{\text{SE}}(\pi) - R(\pi)\} \xrightarrow{d} \mathcal{N}(0, \sigma^2),
    \end{equation*}
    where $\sigma^2 = Var(\varphi_R)$ is the semiparametric efficiency bound of $R(\pi)$, and $\xrightarrow{d}$ means convergence in distribution.
\end{theorem}
% (See Appendix C for proofs) 
Theorem \ref{thm2} (See Appendix A.3 for proofs) establishes the consistency and asymptotic normality of the proposed estimator $\hat{R}_{\text{SE}}(\pi)$. In addition, it shows that 
% By comparing $\hat R_{\text{SE}}(\pi)$ with $\hat R_{\text{Direct}}(\pi)$ and $\hat R_{\text{IPW}}(\pi)$, it can be seen that
 $\hat R_{\text{SE}}(\pi)$ is semiparametric efficient, i.e., it achieves the semiparametric efficiency bound. %(more details can be found in Theorem \ref{thm2}). 
These desired properties hold under the mild condition that the nuisance parameters $\{e(x), s(x), \mu_0(x), \mu_1(x)\}$ are estimated at a rate faster than $n^{-1/4}$. 
These conditions are common in causal inference and can be easily satisfied using a variety of flexible machine learning methods~\cite{chernozhukov2018double,Wu-etal-2024-Harm}. 
% Such conditions are common in causal inference~\cite{chernozhukov2018double} and are easily satisfied using many flexible machine learning methods.   

% . Such conditions are standard in causal inference~\cite{chernozhukov2018double} and can be readily met using various flexible machine learning methods."
% on the convergence rate of the estimated nuisance parameters, specifically 
% These desired properties hold under mild conditions on the convergence rate of the estimated nuisance parameters, as commonly used in causal inference \cite{chernozhukov2018double}. Specifically, these conditions are easily satisfied if the nuisance parameters are estimated at a rate faster than $n^{-1/4}$, which is achievable by many flexible machine learning methods.

%These desired properties hold under mild conditions regarding the convergence rate of the estimated nuisance parameters, which are commonly used in causal inference \cite{chernozhukov2018double}. These conditions are easily satisfied, provided that the nuisance parameters are estimated at the slower rate of $n^{−1/4}$, a criterion achievable by many flexible machine learning methods.

%These desired properties hold under mild conditions on the convergence rate of the estimated nuisance parameters, commonly used in causal inference \cite{chernozhukov2018double}. Specifically, these conditions are easily satisfied if the nuisance parameters are estimated at a rate of $n^{−1/4}$, which is achievable by many flexible machine learning methods.

% ==========================================
\subsection{Learning the Optimal Policy} \label{sec4-2}
After estimating the reward, we now focus on learning the optimal policy. Recall that for a given hypothesis space $\Pi$, the target policy is given as $\pi^{*}(x) =\arg\max_{\pi \in \Pi} R(\pi)$.  
% If $\pi_{0}^{*}(x) \in \Pi$, $ \pi^{*}(x)$
% is equals to $\pi_{0}^{*}(x)$ in Lemma \ref{lemma1}; otherwise, they may not be equal. The difference between them represents the systematic error caused by the limited hypothesis space.
Through optimizing different estimators of reward ${R}(\pi)$, we obtain different estimator of $\pi^{*}$, denoted as $\hat \pi$, defined by  
\begin{equation}
\label{eq6}
\hat \pi(x) = \arg \max_{\pi \in \Pi}\hat { R}(\pi) 
\end{equation}
where $\hat R(\pi)$ can be $ \hat{R}_{\text{Direct}}(\pi)$, $\hat{R}_{\text{IPW}}(\pi)$, or $\hat{R}_{\text{SE}}(\pi)$. 
Algorithm \ref{alg:1} summarizes the procedures for learning $\pi^*$.   
% %is defined in Proposition \ref{prop1}.
% % the learned optimal policy of $\pi^{*}(x)$.

\begin{algorithm}[t]
\caption{Proposed Policy Learning Approach} % Covariate Shift
\label{alg:1}
\textbf{Input}: The source dataset $\mathcal{D}_1$ and the target dataset $\mathcal{D}_0$.  \\
%={}  \{ (X_i,A_i,Y_i,G_i=1): i = 1,...,n_1 \}$, the target dataset $
%\mathcal{D}_0 ={}  \{ (X_i,G_i=0): i = n_1 + 1,...,n_1 + n_0 \}$\\
\textbf{Output}: The learned policy $\hat \pi$. % policy error and welfare change $\Delta W$ % estimated reward $\hat R(\hat \pi)$ and
\begin{algorithmic}[1] 
\STATE\textbf{Stage 1:} Fit models $\hat \mu_1(X)$, $\hat \mu_0(X)$, $\hat e_1(X)$, $\hat s(X)$. % for estimating $\mu_1(X)$, $\mu_0(X)$, $e_1(X)$, $s(X)$. 
% \STATE Define PolicyNet and optimizer
% and linear regression models
% \STATE Define \texttt{PolicyNet} and \texttt{Optimizer}
\STATE\textbf{Stage 2:}
\WHILE{Stop condition is not reached} 
\STATE  Sample a batch of data from $\mathcal{D}_0 \cup \mathcal{D}_1$. 
% \STATE Compute predictions: $\hat \mu_1(X)$, $\hat \mu_0(X)$, $\hat e_1(X)$, $\hat s(X)$
\STATE Minimize the loss -$\hat R(\pi, \hat \mu_0, \hat \mu_0, \hat e_1,\hat s, X, A, Y, G)$ to update $\pi$, using the batch sample. \\
% \texttt{loss $\gets$ -$\hat R(\pi, \hat \mu_0, \hat \mu_0, \hat e_1,\hat s, X, A, Y, G)$} 
% \FOR{epoch \textbf{in} 1 : experiment\_epoch}
%     \FOR{batch \textbf{in} 1 : total\_batches}
%         % \STATE Select batch indices and extract batches
%         \STATE 
%         \STATE Compute loss:\\ \texttt{loss $\gets$ -$\hat R(\hat \mu_0, \hat \mu_0, \hat e_1,\hat s, X, A, Y, G)
%         $}
%         \STATE Perform backpropagation and update PolicyNet weights
%     \ENDFOR
\ENDWHILE
\STATE Return a learned policy. 
% Compute \pw{the estimated reward $\hat R(\hat \pi)$}, policy error \col{XXX} and welfare change $\Delta W$
\end{algorithmic}
\end{algorithm}

% By the definition of $\hat{R}_{\text{Direct}}(\pi)$, it is interesting to note that 
%    \begin{align*}
%       & \arg \max_{\pi} \hat{R}_{\text{Direct}}(\pi) \\
%       ={}& \arg \max_{\pi} \frac{1}{n} \sum_{i = 1}^{n}   \frac{(1-G_i)\pi(X_i)}{1-q}  \Big \{   \hat \mu_1(X_i) - \hat \mu_0(X_i) \Big \} \\
%       ={}&  \begin{cases}1,&\text{if }\hat \tau(X) = \hat \mu_1(X) - \hat \mu_0(X)\geq0\\
%       0,&\text{if }\hat \tau(X) = \hat \mu_1(X) - \hat \mu_0(X) < 0, \end{cases}
%    \end{align*} 
% which implies that the direct method essentially approximates the oracle policy using the plug-in method.  

As discussed in \cite{athey2021policy}, when learning the policy by optimizing the estimated reward, it achieves better generalization performance if the estimated reward is more efficient. Since the estimator $\hat{R}_{\text{SE}}(\pi)$ is the most efficient one under Assumptions \ref{assumption1}-\ref{assumption2}, as shown in Theorem \ref{theo1}, we then focus on exploring its finite sample properties and the learned policy obtained by optimizing it. 
Similar results can also be derived for the direct and IPW methods. 
In finite samples, we allow $\hat \mu_a(X)$, $\hat e_1(X)$, and $\hat s(X)$ to be inaccurate, i.e., they may differ from  $\mu_a(X)$, $e_1(X)$, and $s(X)$.  
% Prposition \ref{prop2} and Theorem \ref{thm2} presents the asymptotic properties of the estimator $\hat{R}_{\text{SE}}(\pi)$. 
% Next, we explore the finite sample property of the learned policy $\hat \pi^*(x)$, analyzing the bias and the generalization error bound.
% However, when all of the models are inaccurate,the bias of MR is given in Theorem 3.

	% with $\hat p_{u,i} > 0$ for all $(u,i)$ pairs, 
The following Proposition \ref{prop3} presents the bias of  $\hat{R}_{\text{SE}}(\pi)$. 
 
\begin{prop}[Bias]  \label{prop3}
    Given the learned $\hat \mu_a(X)$ for $a= 0, 1$, $\hat e_1(X)$, and $\hat s(X)$, then for any given  $\pi$,  the bias of $\hat{R}_{\text{SE}}(\pi)$ is % is given as 
   {\small\begin{align*}
      &  \text{Bias}(\hat{R}_{\text{SE}}(\pi)) ={}  | \mathbb{E}[ \hat{R}_{\text{SE}}(\pi)] - R(\pi) |  \\
     & ={} \Big | \frac 1n \sum_{i=1}^n \Big  [ \frac{\pi(X_i)(\mu_1(X_i) - \hat \mu_1(X_i))}{1-q}\times  \\
{}\quad & \Big \{ \frac{ s(X_i) e_1(X_i)(1 - \hat s(X_i)) - \hat s(X_i) \hat e_1(X_i) (1-s(X_i)) }{ \hat e_1(X_i) \hat s(X_i)} \Big \}       \\
  &+{} \frac{(1-\pi(X_i))(\mu_0(X_i) - \hat \mu_0(X_i))}{1-q} \times \\
 {}&  \Big \{ \frac{ s(X_i) e_0(X_i)(1 - \hat s(X_i)) - \hat s(X_i) \hat e_0(X_i)) (1-s(X_i)) }{ \hat e_0(X_i)) \hat s(X_i)}  \Big \} \Big ] \Big |,
   \end{align*}}where $e_0(X_i) = 1-e_1(X_i)$ and  $\hat e_0(X_i) = 1-\hat e_1(X_i)$.   
\end{prop} 

% (See Appendix D for proofs) % 
From Proposition \ref{prop3} (See Appendix A.4 for proofs), the bias of $\hat{R}_{\text{SE}}(\pi)$ is the product of the estimation errors $\mu_a(X) - \hat \mu_a(X)$) and $s(X)e_a(X)(1-\hat s(X)) - \hat s(X)\hat e_a(X)(1- s(X))$ for $a = 0, 1$. Clearly, when either $\hat \mu_a(X)$ is close to $\mu_a(X)$, or $\hat s(X)$ and $\hat e_1(X)$ are close to $s(X)$ and $e_1(X)$, $\hat{R}_{\text{SE}}(\pi)$ will be close to $R(\pi)$. This further demonstrates the double robustness of $\hat{R}(\pi)$.

Next, we show the generalization error bound (or the regret) of the learned policy. For clarity, we define 
      \begin{align*}
          \hat \pi_{\text{se}}(x) ={}& \arg \max_{\pi \in \Pi} \hat{R}_{\text{SE}}(\pi),
      \end{align*}   
  which is the learned policy by optimizing $\hat{R}_{\text{SE}}(\pi)$.      % the efficient estimator 

% % then for any $\hat \pi \in \Pi$,
\begin{theorem}[Generalization Error Bound]   \label{thm3}
     For any finite hypothesis space $\Pi$, we have that 
     
(i)  with at least probability $1 - \eta$,
\begin{align*} 
  R(\hat \pi_{\text{se}})  
  &\leq \hat R(\hat \pi_\text{se}) + \text{Bias}(\hat{R}_{\text{SE}}(\hat \pi_{se})) + \mathcal{B}(\mathcal{D}_0, \mathcal{D}_1, \eta, \Pi),
\end{align*}
where $\mathcal{B}(\mathcal{D}_0, \mathcal{D}_1, \eta, \Pi)$ equals to 
 \begin{align*}
&\sqrt{ \frac{ \log(2 |\Pi| /\eta) }{ 2 n^{2}  }  \sum_{i=1}^n \frac{ (Y_i - \hat\mu_{A_i}(X_i))^2 (1- \hat s(X_i))^2 }{(1-q)^2 \hat e_{A_i}^2(X_i) \hat s^2(X_i) }},\end{align*} 
with $\hat \mu_{A_i}(X_i) = A_i \hat\mu_1(X_i) + (1-A_i) \hat \mu_0(X_i)$. 
%$\hat \pi_{A_i}(X_i) = A_i \hat \pi(X_i) + (1-A_i)(1- \hat \pi(X_i))$ and \pw{$\hat \mu_{A_i}(X_i) = A_i \hat\mu_1(X_i) + (1-A_i) \hat \mu_0(X_i)$}.  
% \hat \pi_{A_i}^2(X_i)}

%  \pw{if $\pi^*\in \Pi$},
(ii) with at least probability $1 - \eta$, 
\begin{align*}
   R(\hat \pi_{se})
  &\leq R( \pi^*) + \text{Bias}(\hat{R}_{\text{SE}}(\hat \pi_{se})) + \text{Bias}(\hat{R}_{\text{SE}}(\pi^*)) \\
  &\quad+ 2 \mathcal{B}(\mathcal{D}_0, \mathcal{D}_1, \eta, \Pi).
\end{align*}  
\end{theorem}
% thm3
 %Theorem \ref{thm5} gives the {generalization error bounds} of the prediction model trained by minimizing our proposed N-IPS and N-DR estimators.
% Theorem \ref{thm3} holds for any $f_\theta \in \cH$, including $$\hat f_{\theta}^* = \arg \min_{f_\theta \in \cH } \cL_{\text{N-DR}}(\theta, \phi_g|\pi).$$  
% In addition, according to the proof of Theorem \ref{thm4}, both the bias term and the variance term in the generalization bound of  $\cL_{\text{N-DR}}(\theta, \phi_g|\pi)$ are weighted averages of the counterparts in $\cL_{\mathrm{DR}}^\mathrm{N}(\hat \bR |  \boldsymbol{g})$, with a weight of $\pi(\boldsymbol{g})$.

 % where  $\hat e_{u,i}^{*}$ is the prediction error associated with  $\hat \bR^{*}$,  $\hat e_{u,i}^{\dag}$ is the prediction error corresponding to the prediction matrix $\hat \bR^{\dag} = \arg \max_{ \hat \bR^{h} \in \cH } \sum_{(u,i) \in \cD } (e_{u,i} -\hat e_{u,i}^{h})^2 /\hat p_{u,i}^2$. 
 %$\hat \bR^{\dag} = \arg \max_{ \hat \bR^{h} \in \cH } \sum_{(u,i) \in \cD } (  \frac{ e_{u,i} -\hat e_{u,i}^{h}  } { \hat p_{u,i} }  )^{2}$.  
% Theorem \ref{thm3}(a) gives the generalization error bound of $\hat \pi_{se}$.  Theorem \ref{thm3}(b) shows the difference between the generalization risks of the learned policy and the oracle policy.   
% (See Appendix E for proofs)
Theorem \ref{thm3}(i) provides the generalization error bound of the learned policy $\hat \pi_{se}$, and Theorem \ref{thm3}(ii) presents the difference between the generalization risks of the learned policy and the optimal policy $\pi^*$. Note that when $(Y_i - \hat \mu_A(X_i))^2$ are bounded, then $\mathcal{B}(\mathcal{D}_0, \mathcal{D}_1, \eta, \Pi)$ will converge to 0 as the sample size $n$ goes to infinity. 
 Thus, for a sufficiently large sample size $n$,
  if the nuisance parameters are estimated with adequate accuracy, the generalization bound of the learned policy will be approximated well by the estimated reward. Additionally, the generalization bound of the learned policy will be close to that of the optimal policy $\hat \pi^*$. 
 % if the nuisance parameters are estimated with enough accuracy, then the generalization bound of the learned policy is well dominated by the estimated reward $\hat R(\hat \pi_{se})$. In addition, the generalization bound of the learned policy is close to that of the oracle policy.  

% generalize
% ================================================
\subsection{Generalizing Policy to the Entire Domain}
\label{sec4-3}
The approach proposed in Sections 4.1–4.2 is designed to learn the optimal policy in the target domain. In this subsection, we extend the approach to the entire domain, aiming to learn the optimal policy that maximizes $V(\pi)$. 
% Similarly, 
Under Assumption \ref{assumption1}--\ref{assumption2}, the reward $V(\pi)$ is identified as 
\begin{align}
\label{eq6}
V(\pi)
%\mathbb{E}[\pi(X)Y(1) + (1-\pi(X)) Y(0)] \notag \\
= \mathbb{E}[ \pi(X)\mu_{1}(X) + (1-\pi(X)) \mu_{0}(X) ]
\end{align}
Similar to Theorem \ref{theo1}, we present the 
% we employ semiparametric efficiency theory to derive
the efficient influence function and the efficient bound of $V(\pi)$. 
\begin{theorem}[Efficiency Bound of $V(\pi)$] 
\label{theo4} Under Assumptions \ref{assumption1}-\ref{assumption2}, the efficient influence function of  $V(\pi)$ is  
{\small\begin{align*}
& \varphi_{V} ={}\pi(X)\mu_{1}(X) + (1-\pi(X))\mu_{0}(X)  - V(\pi) 
 +   \frac{G}{ s(X) }  \\
{}& \times \left \{ \frac{\pi(X) A (Y - \mu_1(X))}{e_1(X)}+  \frac{(1-\pi(X))(1-A) (Y - \mu_0(X))}{1 - e_1(X)}    \right \}.   
\end{align*}}
The semiparametric efficiency bound of $V(\pi)$ is $\text{Var}(\varphi_{V})$.
\end{theorem}

% Theorem \ref{theo4} (See \col{Appendix}  for proofs) presents the efficient influence function and semiparametric efficiency bound of  $V(\pi)$ under Assumptions \ref{assumption1}-\ref{assumption2}. 
% the efficient influence function of $R(\pi)$,
%which are crucial for constructing efficient estimators of $R(\pi)$.
From Theorem \ref{theo4}, we can construct the semiparametric efficient (SE) estimator of $V(\pi)$, which is given as
\begin{align*}
\hat{V}_{\text{SE}}(\pi) ={}& \frac{1}{n} \sum_{i = 1}^{n} \Bigg[  
\frac{G_i}{\hat{s}(X_i)} \Bigg\{\frac{\pi(X_i) A_i \{Y_i - \hat{\mu}_1(X_i)\}}{\hat{e}_1(X_i)}  \\
+{}&  \frac{(1 - \pi(X_i)) (1-A_i)\{Y_i - \hat{\mu}_0(X_i)\}}{1 - \hat{e}_1(X_i)} \Bigg\} \\
+{}&   \pi(X_i) \hat{\mu}_1(X_i) + (1 - \pi(X_i)) \hat{\mu}_0(X_i) \Bigg]
\end{align*}
Similar to Proposition \ref{prop2} and Theorem \ref{thm2}, it can be shown that $\hat{V}_{\text{SE}}(\pi)$ possesses desirable properties, including double robustness, consistency, asymptotic normality, and semiparametric efficiency under regular conditions.  
% , $\hat V_{\text{SE}}(\pi)$ possesses the desirable properties of double robustness and consistency as $\hat R_{\text{SE}}(\pi)$.
Moreover, following the method described in Section \ref{sec4-2}, we can develop an approach to learn the optimal policy in the entire domain based on the estimated reward $\hat{V}_{\text{SE}}(\pi)$. For brevity, the detailed description is omitted here. 
\begin{table*}[ht]
\centering
\resizebox{1.9\columnwidth}{!}{
\begin{tabular}{c|cccc|ccc|ccc} 
\toprule % 表格顶部粗横线
\multicolumn{1}{c|}{Simulated Dataset} & \multicolumn{4}{c|}{\textsc{Rewards}} & \multicolumn{3}{c|}{\textsc{Policy Error}} & \multicolumn{3}{c}{\textsc{Welfare Changes}} \\
\midrule
METHODS & MEAN & RI & SD & $\Delta\mathrm{E}$ & MEAN & RI & SD & $\Delta\mathrm{W}$ & RI & SD \\
\midrule % 表格中间细横线
\text{Direct (baseline)} & 455.24 & - & \textbf{0.32} & 40.62 & 0.45 & - & \textbf{0.0024} & 215194.26 & - & \textbf{800.41}\\
\text{IPW} & 477.17 & 4.82\% &29.38 & 18.69 & 0.33 & -63.10\% & 0.3165 & 269741.06 & 25.24\% & 41875.37\\
\text{SE} & \textbf{490.22*} & \textbf{7.68\%} & 0.85 & \textbf{5.49*} & \textbf{0.09*} & 
\textbf{78.51\%} & 
0.0062 & \textbf{287185.90*} & \textbf{33.43\%} & 1940.79\\
\midrule % 表格底部粗横线
\midrule % 表格顶部粗横线
\multicolumn{1}{c|}{Real-World Dataset} & \multicolumn{4}{c|}{\textsc{Rewards}} & \multicolumn{3}{c|}{\textsc{Policy Error}} & \multicolumn{3}{c}{\textsc{Welfare Changes}} \\
\midrule
METHODS & MEAN & RI & SD & $\Delta\mathrm{E}$ & MEAN & RI & SD & $\Delta\mathrm{W}$ & RI & SD  \\
\midrule % 表格中间细横线
\text{Direct (baseline)} & 47.76 & - & 0.3076 & 13.18 & 0.50 & - & 0.0083 & 24482.59 & - & 1157.53\\
\text{IPW} & 50.00 & 4.68\% & 1.3002 & 10.95 & 0.42 & 15.07\% & 0.0391 & 28468.45 & 16.28\% & 1971.35\\
\text{SE} & \textbf{60.50*} 
& \textbf{26.66\%} & \textbf{0.0017} & \textbf{0.45*} & \textbf{0.08*} & \textbf{84.35\%} & \textbf{0.0002} & \textbf{47175.56*} & \textbf{92.69\%} & \textbf{814.84}\\
\bottomrule % 表格底部粗横线
\end{tabular}}
\caption{Comparison of estimated rewards, policy errors and welfare changes on the simulated and real-world datasets. The best results are highlighted in bold. RI refers to the relative improvement over the corresponding baseline. SD indicates standard deviation.}
\begin{tablenotes} 
     %\centering
    \footnotesize
    \item  Note: * statistically significant results (p-value $\leq 0.05$) using the paired t-test compared with the baseline. 
    \end{tablenotes}
\label{table2}
\end{table*}

% ===================================
% \subsection{Policy Learning in the Presence of Both Covariate and Concept Shift} \label{sec5-2}

% \bcol{For a simple function class of $\psi_0$ and $\psi_1$, such as linear function class, we can treat XXXX as hyper-parameters, and develop the optimal policy learning approach XXXX.}

\section{Experiments}
We conduct experiments on both simulated datasets and real-world datasets to answer the following questions:
\begin{itemize}
\item \textbf{RQ1}: Does the proposed SE method provide a more accurate estimation of the reward? 
%Does the proposed SE method obtains more accurate estimation of the reward?
% improve the accuracy of estimation compared to the direct and IPW methods?

\item \textbf{RQ2}: Does the proposed method learn better policies? 

%Does our method learn a strategy that is closer to the theoretical optimal?
\item \textbf{RQ3}: How does the degree of covariate shift affect the performance of the proposed method?
%Does the size of the covariate shift have an effect on the performance of our method?

% \item \textbf{RQ4}: How does the degree of concept shift affect the robustness of the proposed method? 

% \bcol{robustness to concept shift}
% \item \textbf{RQ5}: \bcol{policy learning with both covariate and concept shift.} 
\end{itemize}

%In this section, we conduct experiments on both simulated datasets and real-world datasets to compare the performance of three proposed estimation methods with different levels of data utilization.

\subsection{Experiments on Simulated Datasets}

%\subsubsection{Datasets}
\paragraph{Datasets.}   
We generate the source and target datasets in the simulated experiment. The source dataset consists of 512 individuals. For each individual, the covariates $X=(X_{1},X_{2},X_{3})^\top \sim \mathcal{N}(\boldsymbol{\mu_1},\boldsymbol{\Sigma_1})$, where $\boldsymbol{\mu_1} = [10, 3, 7]^\top$ is the mean vector and $\boldsymbol{\Sigma_1} \in \mathbb{R}^{3\times3}$ is the covariance matrix with the $(i,j)$-th element being $2^{-|i-j|}$ for $i,j = 1,2,3$. The treatment $A$ is generated from $Bern(0.5)$, and the potential outcomes are generated as follows: $Y(1) = 15 + 0.4\tilde{X}_{1}\tilde{X}_{2}+0.7\tilde{X}_{3}+ \epsilon, Y(0)= 10 + 0.1\tilde{X}_{1}+0.5\tilde{X}_{2}\tilde{X}_{3} + \epsilon$,   
%Then we simulate the potential outcomes as follows.
%  has elements $\boldsymbol{\Sigma}_{1,i,j}$ defined as
% \begin{align*}
% &Y(1) = 15 + 0.4\tilde{X}_{1}\tilde{X}_{2}+0.7\tilde{X}_{3}+ \epsilon,\\
% &Y(0)= 10 + 0.1\tilde{X}_{1}+0.5\tilde{X}_{2}\tilde{X}_{3} + \epsilon
% \end{align*}
where $\tilde{X}_j=X_j\cdot|X|_j^{0.1}+X_j\cdot|X|_j^{0.3}+X_j\cdot|X|_j^{0.5}$ for $j = 1,2,3$ and $\epsilon \sim \mathcal{N}(0,1)$. 
The observed outcome is $Y = A Y(1) + (1-A)Y(0)$. The variables $X, A, Y$ are available in the source dataset.    
% Note that only one observed outcome $Y = Y(A)$ from one treatment $A$ for each unit is observed. Therefore 
% we have the full information on covariate $X$, treatment $A$, and outcome $Y$ for each individual in the source dataset (G = 1). are sampled from
%has elements $\boldsymbol{\Sigma}_{2,i,j}$ defined as
For the target dataset, we generate 2,048 individuals. The covariates $X \sim\mathcal{N}(\boldsymbol{\mu_2},\boldsymbol{\Sigma_2})$, where the mean vector $\boldsymbol{\mu_2} = [9, 4, 6]^\top$ and the covariance matrix $\boldsymbol{\Sigma_2} \in \mathbb{R}^{3\times3}$ has $(i,j)$-th element given by  
$2^{-|i-j|+1}$ for $i,j = 1,2,3$. Only $X$ is available in the target dataset.  In addition,  
to assess the performance of the learned policy, 
we also generate $\{Y(1), Y(0)\}$ for individuals in the target dataset, using the same method as in the source dataset.  
% \col{What is the test dataset?} 
% in the same way as that in the source dataset.  
% counterfactual results $Y(1), Y(0)$ for each individual are also generated
%. The treatment A is a random vector from $Sig(0.01X_{3}^2)$ and the potential outcomes are generated 
% in the same way as that in the source dataset.

%%%%%%%%%%%可加可不加的
% \subsubsection{Compared methods} We learn the policy using the three estimation methods mentioned above.
% \begin{itemize}
%     \item $\hat R_{direct}$ effectively utilises the information from the target dataset, particularly in the extraction of the relationship between the covariates and outcomes.
%     \item $\hat R_{IPW}$ introduces the propensity score, leveraging full information from the source dataset.
%     \item $\hat R_{SE}$ employs a comprehensive approach to data integration, leveraging both the source dataset and the target dataset. It utilizes the transportability to achieve information transferring and minimize errors.
% \end{itemize}
\paragraph{Compared Methods.} The direct and IPW method. 
$\hat \pi_{\text{direct}} = \arg \max_{\pi \in \Pi} \hat{R}_{\text{Direct}}(\pi)$ 
and $\hat \pi_{\text{ipw}} = \arg \max_{\pi \in \Pi} \hat{R}_{\text{IPW}}(\pi)$.

%  given in Section 4

\paragraph{Evaluation Metrics.}  
The same as ~\cite{kitagawa2018should,li2023trustworthy},  
we adopt the metrics below. % to evaluate the performance.  
%the true reward for the learned policy, policy error, and welfare change to evaluate the performance.  Specifically,
   %\begin{itemize}[leftmargin=*] % the three  s (Direct, IPW, and SE)
%\item

 $\bullet$  To assess the performance of a policy learning method, we calculate the true reward in the target dataset given by  %which is given by 
       $$\hat R(\hat \pi) = n_0^{-1} \sum_{i=1}^n (1-G_i) [ \pi(X_i)Y_i(1) + (1- \pi(X_i))Y_i(0)].$$ 
       % , $\hat{R}(\hat \pi_{\text{ipw}})$, and $\hat{R}(\hat \pi_{\text{se}})$
    Also, we define the regret as   
       $\Delta E = \hat R( \pi_0^* ) - \hat R(\hat \pi)$, 
    representing the difference between the reward induced by the oracle policy and that induced by the learned policy $\hat \pi$. 
       %where $\hat R_{\text{oracle}} = n_0^{-1} \sum_{i=1}^n \lxq{(1-G_i)} [\pi_0^*(X_i)Y_i(1) + (1-\pi_0^*(X_i))Y_i(0)]$  
       % and  is the corresponding estimated reward for each competing method.   
       %To assess the accuracy of the estimation of reward, we compare the estimated values of reward: $\hat{R}_{\text{Direct}}(\hat \pi_{\text{direct}})$, $\hat{R}_{\text{IPW}}(\hat \pi_{\text{ipw}})$, and $\hat{R}_{\text{SE}}(\hat \pi_{\text{se}})$. Also, we define the regret as    
       
      $\bullet$   To evaluate the accuracy of the learned policy $\hat{\pi}$, we define the policy error as 
       $$n_0^{-1} \sum_{i=1}^n(1-G_i)|\pi_0^*(X_i)-\hat{\pi}(X_i)|^2,$$
        the mean square errors between the oracle policy $\pi_0^*$ and $\hat{\pi}$. 
       % $\hat{\pi}$
      %  oracle policy $\hat \pi_{Direct}$, $\hat \pi_{IPW}$, $\hat \pi_{SE}$ learned by three estimation methods. The policy error is defined as 
  
         % Finally we give a comparison of welfare changes. Similar as \cite{kitagawa2018should,li2023trustworthy},
      % is defined 
   $\bullet$  %In addition, 
We define welfare change as
$$\Delta W=\sum_{i=1}^{n}[(Y_{i}(1)-Y_{i}(0))\hat{\pi}(X_{i})],$$ representing the difference between the total rewards induced by $\hat \pi$ and the null policy $\pi \equiv 0$. % the total reward from the
%which is the difference between the total reward from $\hat \pi$ and the total reward from the null policy $\pi \equiv 0$.  
A policy learning method is better when it yields a larger true reward, smaller regret, lower policy error, and greater welfare change. 

\begin{figure*}[h!]
\centering
\begin{subfigure}[b]{0.3\textwidth}
    \centering
    \includegraphics[width=\textwidth]{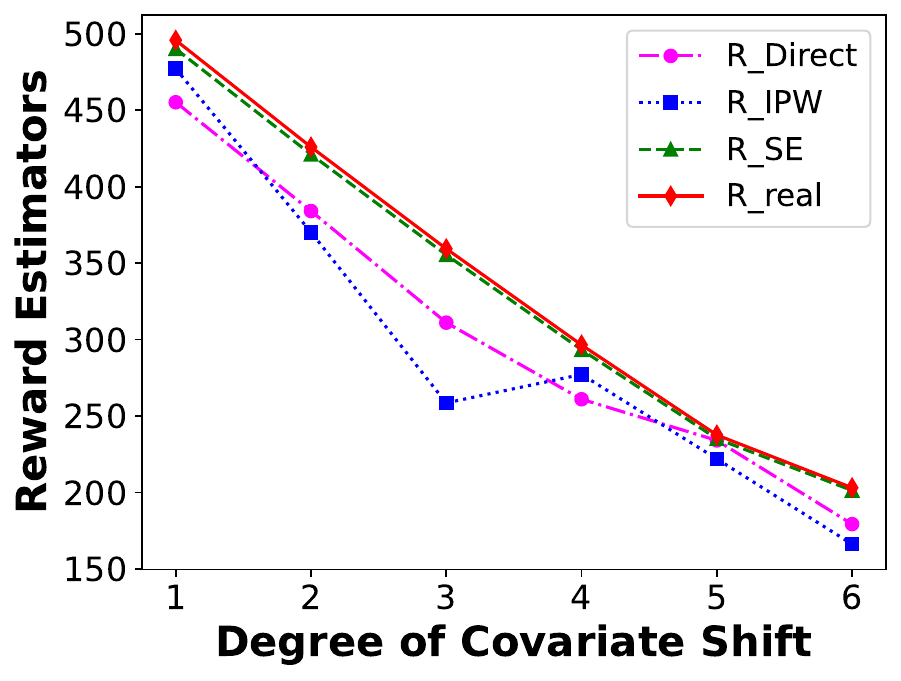}
    \caption{Reward estimators }
    \label{fig:sub1}
\end{subfigure}
\begin{subfigure}[b]{0.3\textwidth}
    \centering
    \includegraphics[width=\textwidth]{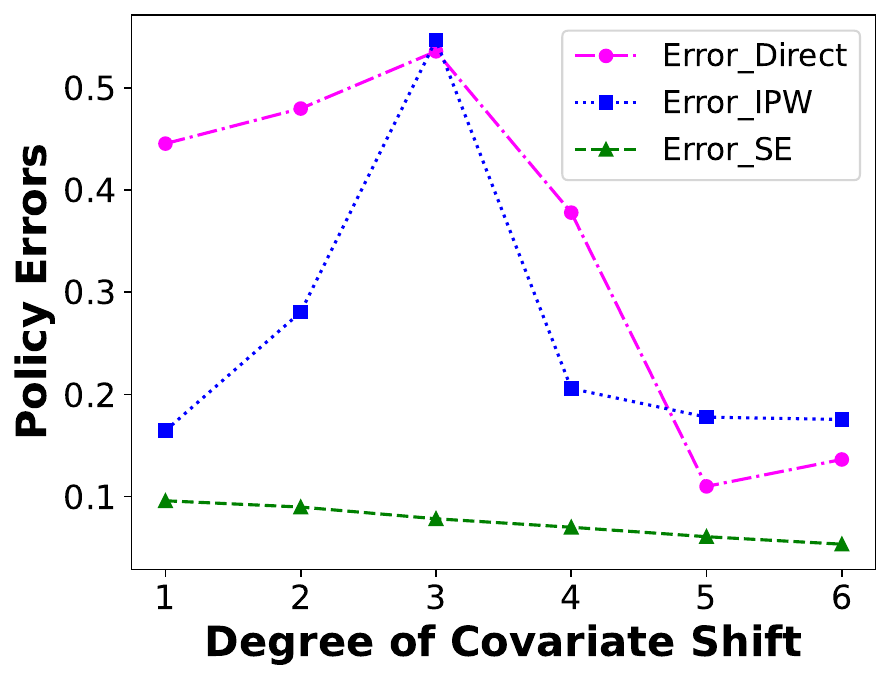}
    \caption{Policy errors}
    \label{fig:sub1}
\end{subfigure}
\begin{subfigure}[b]{0.3\textwidth}
    \centering
    \includegraphics[width=\textwidth]{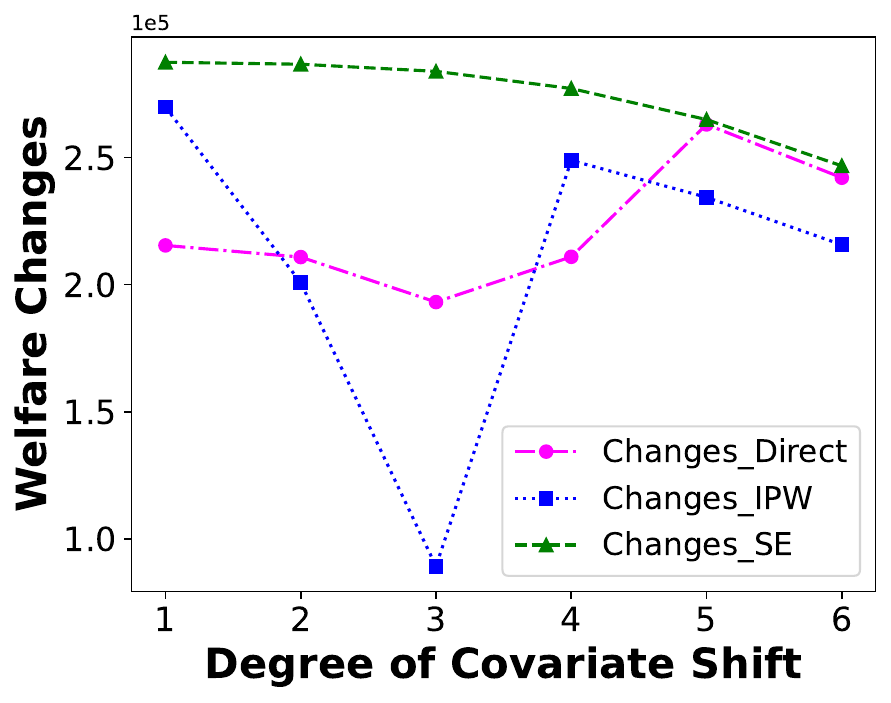}
    \caption{Welfare changes}
    \label{fig:sub2}
\end{subfigure}
\caption{Comparison of three methods with different means of covariates in the target dataset}
\label{fig1}
\end{figure*}

\begin{figure*}[h!]
\centering
\begin{subfigure}[b]{0.3\textwidth}
    \centering
    \includegraphics[width=\textwidth]{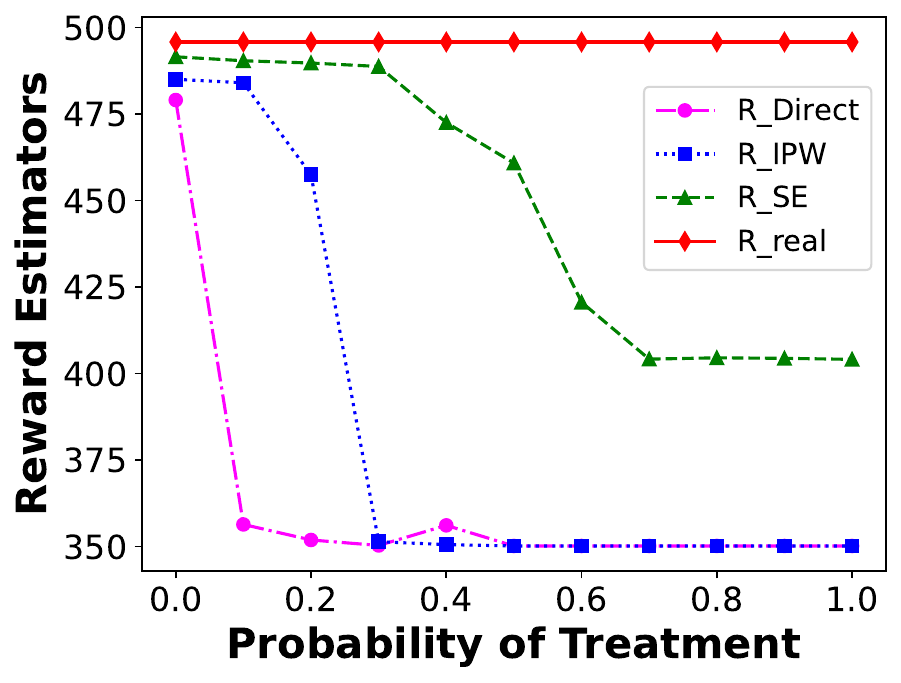}
    \caption{Reward estimators }
    \label{fig:sub1}
\end{subfigure}
\begin{subfigure}[b]{0.3\textwidth}
    \centering
    \includegraphics[width=\textwidth]{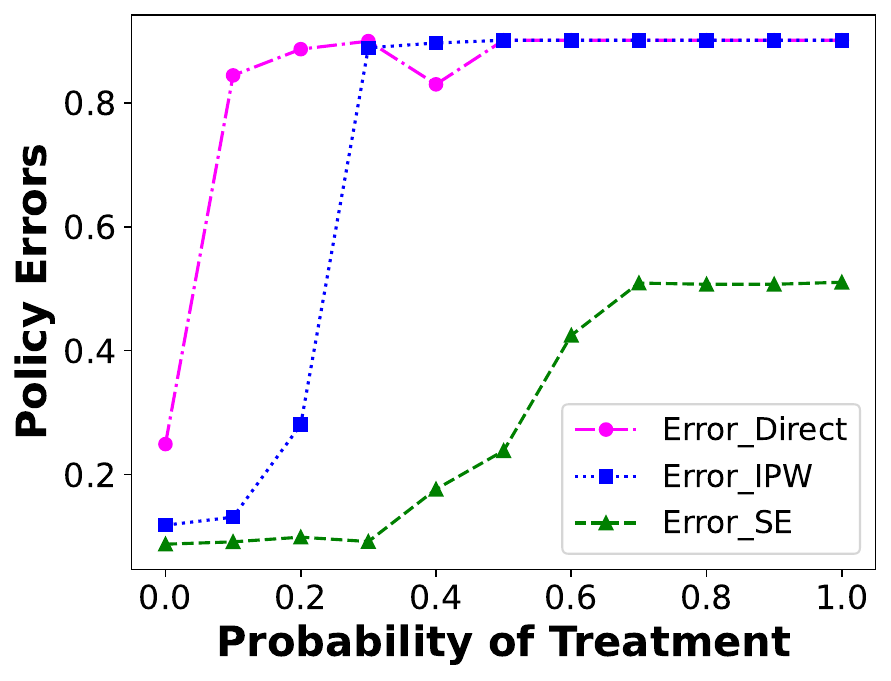}
    \caption{Policy errors}
    \label{fig:sub1}
\end{subfigure}
\begin{subfigure}[b]{0.3\textwidth}
    \centering
    \includegraphics[width=\textwidth]{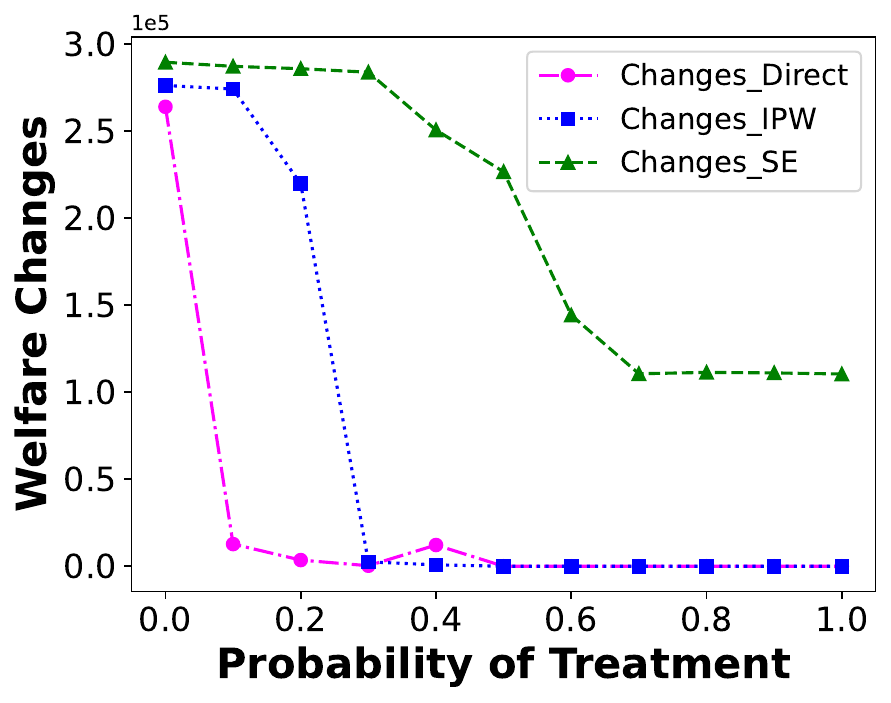}
    \caption{Welfare changes}
    \label{fig:sub2}
\end{subfigure}
\caption{Comparison of three methods with different treatments in the target dataset}
\label{fig3}
\end{figure*}
% \begin{figure*}[h!]
% \centering
% \begin{subfigure}[b]{0.3\textwidth}
%     \centering
%     \includegraphics[width=\textwidth]{plot7.pdf}
%     \caption{Reward estimators }
%     \label{fig:sub1}
% \end{subfigure}
% \begin{subfigure}[b]{0.3\textwidth}
%     \centering
%     \includegraphics[width=\textwidth]{plot8.pdf}
%     \caption{Policy errors}
%     \label{fig:sub1}
% \end{subfigure}
% \begin{subfigure}[b]{0.3\textwidth}
%     \centering
%     \includegraphics[width=\textwidth]{plot9.pdf}
%     \caption{Welfare changes}
%     \label{fig:sub2}
% \end{subfigure}
% \vskip -8pt 
% \caption{Comparison of three methods with different concept shift parameters in the target dataset}
% \label{fig2}
% \vskip -12pt 
% \end{figure*}

\paragraph{Performance Comparison (RQ1 and RQ2).}  
% 目的是学习最大化reward的最优政策
% 考虑三种估计方法，direct，IPS，SE
% 训练50次，计算3个R的均值作为REWARD，标准差std，和real的差距如表所示。（另外评估学习的pi）We
%report the rewards, the changes in welfare, and policy errors with different estimation methods.
%  in simulated datasets
We average 50 independent trials of policy learning and Table \ref{table2} (top panel) reports the average rewards $\hat R(\hat \pi)$, regrets $\Delta E$, policy errors, and the welfare changes for different policy learning methods. From it, we have the following observations: 

% (top panel)
% The objective of this study is to learn the optimal policy based on maximising the reward in target dataset, utilising the three estimation methods previously outlined and comparing them. In order to facilitate the assessment of the performance of the optimal policies derived from different estimation methods, the analysis is conducted from two distinct perspectives. 
%  and is closest to $\hat R_{real}$
% Compared to the direct and IPW methods, 
% That is, $\hat R_{SE}$ gives better estimation accuracy.  
 $\bullet$ The proposed SE method achieves the highest reward, smallest regret, lowest policy error, and largest welfare change. Compared to the Direct method (baseline), these improvements are substantial---up to 7.68\% in reward, 78.51\% in policy error, and 33.43\% in welfare change---demonstrating the superiority of the SE method. 
 $\bullet$ The standard deviations of the Direct and SE methods are significantly smaller than that of the IPW method,  indicating the instability (large variance) of the IPW method. 
% The proposed SE method has the smallest policy error. The policy error for the SE method is below 0.1, while the errors for the Direct and IPW methods exceed 0.3. 

% Additionally, the SD of the policy errors for the SE and Direct methods is smaller than that of the IPW method.  
 % giving a better policy estimation. 
% On the other hand, 
 % suggesting that the IPW method suffers in this aspect of the policy learning due to its high sensitivity to $e_1(x)$ and $s(x)$.
% \item The proposed SE method has the largest welfare change compared to the Direct and IPW methods.  
% The estimator from our proposed method are significantly higher than that from the other two methods.

In summary, the SE method outperforms the competing Direct and IPW methods, owing to its desirable properties, such as double robustness and semiparametric efficiency. 
% theoretical 

\paragraph{Effect of varying the degree of covariate shift (RQ3).}   
We evaluate the robustness of three policy learning approaches (Direct, IPW, SE) by varying the degree of covariate shift between the source and target datasets. 
The degree of covariate shift is measured by the Chebyshev distance between the two mean vectors $\boldsymbol{\mu_1}$ and $\boldsymbol{\mu_2}$ in source and target datasets. The Chebyshev distance is defined as $d(\boldsymbol{\mu_1}, \boldsymbol{\mu_2}) = \max_{j} |\mu_{1,j} -\mu_{2,j}|$, where $\mu_{1,j}$ and $\mu_{2,j}$ are the $j$-th elements of  $\boldsymbol{\mu_1}$ and $\boldsymbol{\mu_2}$, respectively.
% We set  
%where $d(\mathbf{\mu_2}, \mathbf{\mu^*}) = \max_{i=1,2,3} |\mu_{2, i} - \mu^*_i|$, which represents the maximum value of the difference between the new mean $\mu^*$ and initial mean $\mu_2$ of the covariates in each dimension. 
% and observing the corresponding changes in model performance.
Three metrics are used to measure performance: true reward induced by the learned policy, policy error, and welfare change. Figure \ref{fig1} illustrates how these metrics vary with increasing covariate shift, showing that the proposed SE method remains more stable across different levels of covariate shift and consistently achieves superior performance. 

%\vspace{-1pt}
% \smallskip
% \noindent \textbf{Effect of varying the degree of concept shift (RQ4).} 
% We also evaluate the robustness of the proposed methods by varying the degree of concept shift between the source and target datasets. The degree of covariate shift is measured by concept shift parameters $\epsilon_1$ and $\epsilon_0$. To simplify, we let $\epsilon_1 = \epsilon_1 = 0.1, 0.2, \ldots, 0.9, 1$. Figure \ref{fig2} demonstrates the variation of these metrics as concept shift decreases, highlighting the greater performance of the proposed SE method across varying levels of concept shift. 

\paragraph{Effect of varying probability of treatment.} 
 We further evaluate the robustness of the proposed methods by varying the generation mechanism of treatments in the source dataset.
We generate $A$ in the source dataset from $Bern(\sigma(-\beta \tilde{X}_2))$, where $\sigma(\cdot)$ is the sigmoid function, and $\beta = 0, 0.1, \dots, 0.9, 1$. 
The probability of receiving the treatment is 0.5 when $\beta=0$. As $\beta$ increases, the probability of an individual receiving the treatment decreases (since $\tilde{X}_2$ is always greater than 0 in our setting). Figure \ref{fig3} displays how the evaluation metrics change as the proportion of individuals receiving treatment decreases, demonstrating the robustness, stability, and overall superiority of the proposed SE method.

\subsection{Experiments on Real-World Datasets}
\paragraph{Datasets.}The Communities and Crime dataset~\cite{misc_communities_and_crime_183} comprises 1994 records from communities in the United States, which combines socio-economic data from the 1990 US Census, law enforcement data from the 1990 US LEMAS survey, and crime data from the 1995 FBI UCR. Each record includes 127 covariates, including location information (such as state and county), economic factors (such as perCapInc and HousVacant) and demographics (such as PopDens and PctBSorMore).  
We use records from communities in New Jersey as the source dataset and records from communities in other states as the target dataset. 
In addition to using the information from covariates, we simulated the treatment $A$ and the potential outcomes $Y(1), Y(0)$. See Appendix D for the detailed data generation process. % details of
\paragraph{Performance Comparison.}We also average over 50 independent trials of policy learning and  Table \ref{table2} (bottom panel) reports the associated results. 
% the average true rewards $\hat R(\hat \pi)$, regrets $\Delta E$, policy errors, and the welfare changes for different policy learning methods. 
From it, we have the following observations: (1) The proposed SE method outperforms the Direct and IPW methods across all evaluation metrics; (2) Compared to the baseline, the SE method shows substantial improvements—up to 26.66\% in reward, 84.35\% in policy error, and 92.69\% in welfare change. 
%(3) The standard deviations of the SE method are significantly smaller than those of the Direct and IPW methods. 
These observations further demonstrate the SE method's superiority.  
% indicating the instability (large variance) of the IPW method.  
% \begin{itemize}
% \item The proposed SE method achieves the highest reward, smallest regret, lowest policy error, and largest welfare change. Compared to the Direct method (baseline), these improvements are substantial---up to 26.66\% in reward, 84.35\% in policy error, and 92.69\% in welfare change---demonstrating the superiority of the SE method. 
% \item  The standard deviations of the SE method are significantly smaller than that of the Direct and IPW methods,  indicating the instability (large variance) of the IPW method. 
% \end{itemize}
% In summary, the SE method outperforms the competing Direct and IPW methods, owing to its desirable properties, such as double robustness and semiparametric efficiency. 
\section{Related Work}
\paragraph{Policy Learning.} Policy learning seeks to identify which individuals should receive treatment to maximize the reward according to their covariates~\cite{murphy2003optimal},  with wide-ranging applications in fields such as precision medicine~  \cite{bertsimas2017personalized,kitagawa2018should,Kosorok+Laber:2019}, 
reinforcement learning~\cite{liu2021policy,kwan2023survey}, and recommender systems~\cite{ma2020off,chen2021knowledge,li2023balancing}.     
Nevertheless, most policy learning approaches rely solely on a single labeled dataset. In scenarios where labeled data are difficult to obtain, these methods often struggle with external validity and generalizability. To address this issue, a straightforward strategy is to combine labeled and unlabeled data, utilizing techniques like transfer learning and semi-supervised learning ~\cite{kora2022transfer,huynh2022semi}. However, while transfer and semi-supervised learning methods are well-developed for prediction tasks, the challenge of learning optimal policies by leveraging both labeled and unlabeled data remains largely unexplored~\cite{uehara2020off}.  
\paragraph{Causal Effects Generalizability.} Recently, there has been growing research interest in integrating information from multiple data sources for causal inference~\cite{hartman2015sample,lodi2019effect,colnet2024causal,Wu-etal-2025-Compare,kallus2024role,wu-mao2025}. However, heterogeneity in data distribution across these sources presents a significant challenge. Different heterogeneity in data distribution necessitates tailored techniques, including specific assumptions~\cite{hotz2005predicting,kern2016assessing,li2023improving,hu2023longterm,Wu-etal-ShortLong,yang2024learning} and structural causal models (SCM)~\cite{pearl1995causal,correa2018generalized,tikka2019causal}. While data integration has gained considerable theoretical attention, its extension to policy learning remains underexplored. In this article, we investigate how to learn the optimal policy in a target dataset by leveraging information from a source dataset.

\section{Conclusion}
In this article, we propose a principled policy learning method from a causal perspective. We detail the identifiability assumptions for the reward, derive its efficient influence function, and develop a doubly robust estimator, which is consistent, asymptotically normal, and achieves semiparametric efficiency. We also provide a generalization error bound for the learned optimal policy. 
% Moreover, we introduce a novel sensitivity analysis method to assess the robustness of the proposed policy learning approach under both covariate and concept shifts. 
Extensive experiments confirm the effectiveness and reliability of our proposed method, demonstrating both theoretical and practical advantages. 
%A limitation is that the method requires $0 < s(X) < 1$, which may not hold when covariate distributions between source and target datasets vary substantially. Future research should explore relaxing this assumption to improve the method's applicability in more diverse scenarios.

\section*{Acknowledgments}
This work was supported by 
the National Key Research and Development Program (No. 2024YFD2100700), 
 the National Natural Science Foundation of China (No. 12301370), % \col{Project of JianHua Guo}, 
the BTBU Digital Business Platform Project by BMEC, 
 the Beijing Key Laboratory of Applied Statistics and Digital Regulation, and the gift funding from ByteDance Research.

\newpage 
\bibliographystyle{named}
\bibliography{ijcai25}

\end{document}